\pdfoutput=1
\documentclass[11pt]{article}

\usepackage{acl}

\usepackage{times}
\usepackage{latexsym}
\usepackage{makecell}

\usepackage[T1]{fontenc}
\usepackage[utf8]{inputenc}

\usepackage{microtype}

\usepackage{xcolor}
\definecolor{thedarkblue}{RGB}{0,0,120} %104} % 180
\definecolor{mydarkblue}{rgb}{0,0.08,0.45} %ICML dark blue
\definecolor{darkblue}{rgb}{0,0.08,180}
\colorlet{TufteRed}{red!80!black}
\definecolor{theblue}{RGB}{0,0,180}
\colorlet{thered}{TufteRed}
      
\usepackage{microtype}
\usepackage{balance}

\usepackage{booktabs}
\usepackage{tabularx}

\usepackage{amsmath,amssymb,amsthm}

\newcommand{\eat}[1]{\ignorespaces}
\usepackage{comment}

\newcommand{\journal}[1]{} % text that is too long for short conf paper, but would be good for journal

\usepackage{tikz}
\usepackage{verbatim}
\usetikzlibrary{arrows}
\usetikzlibrary{shapes,snakes}
\usetikzlibrary{decorations.pathmorphing} % noisy shapes
\usetikzlibrary{fit}					% fitting shapes to coordinates
\usetikzlibrary{backgrounds}	% drawing the background after the foreground

\usepackage{ragged2e}
\usepackage{multirow}
\usepackage{microtype}
\usepackage{balance}
\usepackage{setspace}

\graphicspath{{./}{./graphics/}}
\newcolumntype{H}{>{\setbox0=\hbox\bgroup}c<{\egroup}@{}}

\newcolumntype{R}[1]{>{\RaggedLeft\arraybackslash}} %p{#1}}
\newcolumntype{L}[1]{>{\RaggedRight\arraybackslash}} %p{#1}}

\newcommand{\eg}{\emph{e.g.}}

\AtBeginEnvironment{pmatrix}{\setlength{\arraycolsep}{2pt}}

\DeclareMathOperator{\hugeE}{\mbox{\huge\raise-0.3ex\hbox{E}}}
\DeclareMathOperator{\p}{\mathbb{P}}
\DeclareMathOperator{\hugep}{\mbox{\huge\raise-0.3ex\hbox{$\p$}}}

\usepackage{subfigure}
\usepackage{nicefrac}

\DeclareMathAlphabet{\mathbcal}{OMS}{cmsy}{b}{n}
\usepackage{amsmath}
\usepackage{mathrsfs}
\usepackage{comment}

\usepackage{bm}
\usepackage{bbm}
\usepackage{amssymb}
\usepackage{enumitem}
\usepackage{paralist}
\usepackage{graphicx}

\title{Personalized Multimodal Large Language Models: A Survey}

\author{%
Junda Wu\textsuperscript{\rm 1} \quad Hanjia Lyu\textsuperscript{\rm 2} \quad Yu Xia\textsuperscript{\rm 1} \quad Zhehao Zhang\textsuperscript{\rm 3} \quad Joe Barrow\textsuperscript{\rm 4} \\ 
\textbf{Ishita Kumar\textsuperscript{\rm 5} \quad Mehrnoosh Mirtaheri\textsuperscript{\rm 6} \quad Hongjie Chen\textsuperscript{\rm 7} \quad Ryan A. Rossi\textsuperscript{\rm 4} \quad Franck Dernoncourt\textsuperscript{\rm 4}} \\
\textbf{Tong Yu\textsuperscript{\rm 4} \quad Ruiyi Zhang\textsuperscript{\rm 4} \quad Jiuxiang Gu\textsuperscript{\rm 4} \quad Nesreen K. Ahmed\textsuperscript{\rm 8} \quad Yu Wang\textsuperscript{\rm 9} \quad Xiang Chen\textsuperscript{\rm 4}} \\
 \textbf{Hanieh Deilamsalehy\textsuperscript{\rm 4} \quad Namyong Park\textsuperscript{\rm 10} \quad Sungchul Kim\textsuperscript{\rm 4} \quad Huanrui Yang\textsuperscript{\rm 11} \quad Subrata Mitra\textsuperscript{\rm 4}} \\
 \vspace{3pt}
\textbf{Zhengmian Hu\textsuperscript{\rm 4} 
\quad Nedim Lipka\textsuperscript{\rm 4} 
\quad Dang Nguyen\textsuperscript{\rm 12}
\quad Yue Zhao\textsuperscript{\rm 6}
\quad Jiebo Luo\textsuperscript{\rm 2} \quad Julian McAuley\textsuperscript{\rm 1}}  \\
\normalsize\textsuperscript{\rm 1}University of California, San Diego \quad \textsuperscript{\rm 2}University of Rochester \quad
\textsuperscript{\rm 3}Dartmouth College \quad 
\textsuperscript{\rm 4}Adobe Research \\
\normalsize\textsuperscript{\rm 5}University of Massachusetts at Amherst \quad
\textsuperscript{\rm 6}University of Southern California \quad
\textsuperscript{\rm 7}Virginia Tech \\
\normalsize\textsuperscript{\rm 8}Cisco Research \quad 
\textsuperscript{\rm 9}University of Oregon \quad
\textsuperscript{\rm 10}Meta AI \quad 
\textsuperscript{\rm 11}University of Arizona \quad
\textsuperscript{\rm 12}University of Maryland\\
}

\begin{document}
\maketitle

\begin{abstract}
Multimodal Large Language Models (MLLMs) have become increasingly important due to their state-of-the-art performance and ability to integrate multiple data modalities, such as text, images, and audio, to perform complex tasks with high accuracy. 
This paper presents a comprehensive survey on personalized multimodal large language models, focusing on their architecture, training methods, and applications. 
We propose an intuitive taxonomy for categorizing the techniques used to personalize MLLMs to individual users, and discuss the techniques accordingly.
Furthermore, we discuss how such techniques can be combined or adapted when appropriate, highlighting their advantages and underlying rationale.
We also provide a succinct summary of personalization tasks investigated in existing research, along with the evaluation metrics commonly used.
Additionally, we summarize the datasets that are useful for benchmarking personalized MLLMs.
Finally, we outline critical open challenges.
This survey aims to serve as a valuable resource for researchers and practitioners seeking to understand and advance the development of personalized multimodal large language models.
\end{abstract}

\section{Introduction}

Multimodal Large Language Models (MLLMs)\footnote{Note MLLM is used synonymously with Large Multimodal Model (LMM), and Large Vision Language Model (LVLM).}
have recently become important for generating and reasoning with diverse types of complex data such as text, images, and audio~\cite{yang2023dawn}.
These models that process, generate, and combine information across modalities have found many applications such as healthcare~\cite{lu2024multimodal,alsaad2024multimodal}, recommendation~\cite{lyu2024x,tian2024mmrec}, autonomous vehicles~\cite{cui2024survey,chen2024driving}.
However, to further enhance the performance and utility of these models, personalization plays a crucial role, enabling them to adapt more effectively to a user's specific preferences, context, and needs~\cite{chen2024large}.
Personalization offers an improved user experience by saving time and increasing accuracy, for instance, by generating content that is more closely aligned with the user's interests.

Personalization in multimodal large language models comes with its own set of unique challenges.
In particular, 
in order to create personalized experiences, it is essential to have individual-level data from users. In multimodal scenarios, this requires data that spans multiple modalities. For instance a user might generate an image from a text prompt and then provide feedback, such as a thumbs-up or thumbs-down. In this case, we have two modalities—text and image—along with implicit feedback on the text prompt and explicit feedback on the generated image in the form of a like or dislike.

\definecolor{googleblue}{HTML}{4285F4}
\definecolor{googlered}{HTML}{DB4437}
\definecolor{googlepurple}{HTML}{A142F4} % New purple color
\definecolor{googlegreen}{HTML}{0F9D58}

\begin{table*}[h]
\centering
\small
\begin{tabular}{cr l}
\toprule
\textbf{Category} & \textbf{General Mechanism} & \textbf{Example Models and Methods} \\ 
\midrule

\multirow{3}{*}{\textcolor{googlepurple}{\textbf{\makecell{Personalized MLLM\\Text Generation}}}} 
& Instruction (Sec.~\ref{sec:text-gen-instruction}) & CGSMP~\cite{yong2023cgsmp}, ModICT~\cite{li2024multimodal}\\ 
\multirow{4}{*}{(\textbf{Section~\ref{sec:text-gen}})}
& Alignment (Sec.~\ref{sec:text-gen-alignment}) & MPDialog \cite{agrawal2023multimodal}, Athena 3.0~\cite{fan2023athena}\\ 
& Generation (Sec.~\ref{sec:text-gen-generation}) & \citet{wu2024understanding}, PTSCG~\cite{wang2024personalized}\\ 
& Fine-tuning (Sec.~\ref{sec:text-finetune}) & \citet{wang2023user}, PVIT~\cite{pi2024personalized} \\ 
\midrule

\multirow{3}{*}{\textcolor{googlered}{\textbf{\makecell{Personalized MLLM\\Image Generation}}}} & Instruction (Sec.~\ref{sec:image-gen-instruction}) & MuDI~\cite{jang2024identity}, \citet{zhong2024user}\\ 
\multirow{4}{*}{(\textbf{Section~\ref{sec:image-gen}})}
& Alignment (Sec.~\ref{sec:image-gen-alignment}) & $\lambda$-ECLIPSE~\cite{patel2024lambda}, MoMA~\cite{song2024moma}\\ 
& Generation (Sec.~\ref{sec:image-gen-generation}) & Layout-and-Retouch~\cite{kim2024layout}, Instantbooth~\cite{shi2024instantbooth} \\ 
& Fine-tuning (Sec.~\ref{sec:image-gen-finetune})  &  MS-Diffusion~\cite{wang2024ms}, UNIMO-G~\cite{li2024unimo} \\ 
\midrule

% sec:rec
\multirow{3}{*}{\textcolor{googleblue}{\textbf{\makecell{Personalized MLLM\\Recommendation}}}} 
& Instruction (Sec.~\ref{sec:rec-instruction}) &  InteraRec~\cite{karra2024interarec}, X-Reflect~\cite{lyu2024x}\\ 
\multirow{4}{*}{(\textbf{Section~\ref{sec:rec}})}
& Alignment (Sec.~\ref{sec:rec-alignment}) & PMG~\cite{shen2024pmg}, MMREC~\cite{tian2024mmrec}\\ 
& Generation (Sec.~\ref{sec:rec-generation}) & RA-Rec~\cite{yu2024ra},\citet{wei2024towards}\\ 
& Fine-tuning (Sec.~\ref{sec:rec-finetune}) & GPT4Rec\cite{zhang2024gpt4rec},MMSSL \cite{wei2023multi}  \\ 
\midrule

\multirow{3}{*}{\textcolor{googlegreen}{\textbf{\makecell{Personalized MLLM\\Retrieval}}}} & Instruction (Sec.~\ref{sec:ret-instruction}) & ConCon-Chi \cite{Rosasco_2024_CVPR}, Med-PMC \cite{liu2024med}\\ 
\multirow{4}{*}{(\textbf{Section~\ref{sec:ret}})}
& Alignment (Sec.~\ref{sec:ret-alignment}) & AlignBot \cite{chen2024alignbot},  \cite{xu2024align}\\ 
& Generation (Sec.~\ref{sec:ret-generation}) & \citet{ye2024contemporary},Yo'LLaVA~\cite{nguyen2024yo} \\ 
& Fine-tuning (Sec.~\ref{sec:ret-finetune}) & FedPAM \cite{feng2024fedpam}, VITR \cite{gong2023vitr}\\

\bottomrule
\end{tabular}
\caption{Overview of Techniques for Personalized Multimodal Large Language Models (Sections~\ref{sec:text-gen}-\ref{sec:ret}).}
\label{tab:overview-techniques}
\end{table*}

We propose an intuitive and symmetric taxonomy for the techniques used for personalized multimodal large language models, 
including 
text generation (Section~\ref{sec:text-gen}), 
image generation (Section~\ref{sec:image-gen}),
recommendation (Section~\ref{sec:rec}), 
and retrieval (Section~\ref{sec:ret}).
Each category classifies techniques based on how they follow multimodal instructions, enable multimodal alignment, 
generate personalized responses, or incorporate personalization through model fine-tuning.
We illustrate an overview of techniques for personalized multimodal models in Table~\ref{tab:overview-techniques}.
In parallel to the discussion of techniques, we also summarize various applications in personalized MLLMs (Appendix~\ref{sec:applications}).

\medskip\noindent\textbf{Summary of Main Contributions.} 
The key contributions of this work are as follows:

\begin{compactitem}
    \item A comprehensive survey of existing work for personalized MLLMs.
    We also survey the problem settings, evaluation metrics, and datasets used in the literature.

    \item We introduce a few intuitive taxonomies for personalization in MLLMs and survey existing work using these taxonomies.
    \item Key open problems and challenges are identified and discussed. These problems are important for future work to address in this rapidly growing but vitaly important field.
\end{compactitem}

\medskip\noindent\textbf{Scope of the survey.} 
In this survey, we focus entirely on recent work that leverages \textbf{multimodal large language models} (MLLMs) to generate \textbf{personalized} text, images, audio, or other modalities.
We consider techniques for personalization that elicit and incorporate \emph{user preferences} when generating multimodal outputs.
To study these techniques, we decompose works across the following three dimensions:
\begin{compactitem}
    \item the \emph{modality} of the content being generated (\eg, text, images, audio, or other);
    \item the \emph{personalization technique} being employed (\eg, prompt-based, prefix tuning, finetuning/adapters);
    \item the \emph{application} of the personalized MLLM (\eg, chat/assistant, recommendation systems, retrieval, classification, image generation, text generation).
\end{compactitem}

\section{Overview of Challenges and Techniques}
Personalization in multimodal large language models presents several significant challenges, due to the complexity of combining diverse types of data, extracting relevant information, and delivering user-specific insights. To tackle these challenges, 
researchers have
introduced techniques such as multimodal instruction, alignment, and generation.

\subsection{Integration of Heterogeneous Data}

Multimodal large language models need to combine information from various modalities, such as text, images, audio, video, and user engagement~\cite{wei2024towards,xu2024align}. Each modality has distinct characteristics and may convey different types of information. For example, text might describe a product, while an image conveys its visual appearance. Integrating these heterogeneous data types is challenging 
because they require different encoding methods, processing pipelines, and alignment strategies. Misalignment or incomplete fusion~\cite{lyu2024x,zhou2023exploring} can lead to inconsistent or inaccurate user preferences being captured, thus reducing the effectiveness of personalized recommendations. Multimodal alignment can help resolve inconsistencies that may arise from mismatched modalities.

\subsection{Data Noise and Redundancy}
Different modalities often include noisy, redundant, or irrelevant information~\cite{liu2024rec,lyu2022understanding}. For example, images of the same product in e-commerce platforms may have varying quality or redundant features, while textual descriptions might include unnecessary details. Extracting meaningful insights from such noisy data is challenging because the model needs to filter out irrelevant content without losing important context. This process becomes even more difficult when handling large amounts of data, as the noise accumulates and complicates the extraction of relevant user preferences. Multimodal instruction can help filter out noisy or redundant data by guiding the model to focus on the most relevant modalities and inputs for each user. By directing the model’s attention to key features in user interactions, this method reduces the impact of irrelevant or repetitive information, ensuring that the generated outputs are more meaningful and concise.

\subsection{Granular Understanding of Multimodal Data}
Text-based LLMs are adept at processing linguistic information such as item descriptions~\cite{lyu2024llm}, and some approaches seek to transform non-textual data into the text space~\cite{ye2024harnessing}. However, visual inputs often contain nuances—such as color, texture, and context—that are difficult to capture with language alone~\cite{shen2024pmg}. For instance, subtle preferences in fashion, home decor, or art may be driven by visual factors that are abstract or subjective. Multimodal LLMs may struggle to extract these fine-grained visual details and relate them meaningfully to textual descriptions, leading to a loss of personalization depth. Multimodal alignment facilitates a mre granular understanding by ensuring that the relationships between different modalities are preserved.

\subsection{Scalability and Efficiency}
As the volume of multimodal engagement grows, so do the computational demands for processing and personalizing recommendations. Models need to handle a large number of user interactions across various modalities in real-time environments~\cite{ye2024harnessing,shen2023scalable}, such as social media platforms or e-commerce sites. This necessitates advanced resource allocation strategies, as multimodal large language models often require significant GPU or TPU resources to process images, videos, or audio in parallel with text.

\subsection{Capturing Diverse and Dynamic User Preferences}
Users interact with multimodal content in diverse ways, and their preferences can evolve over time~\cite{rafailidis2017preference}. Accurately capturing these preferences across modalities is challenging because different data types might signal conflicting or evolving interests. For instance, a user’s engagement with both product reviews and product images may shift over time, requiring the model to adapt its understanding of their preferences in real-time. Additionally, the model needs to continuously update its understanding to reflect new patterns of user behavior.

\section{Personalized MLLM Text Generation}
\label{sec:text-gen}

\subsection{Personalized Multimodal Instruction}\label{sec:text-gen-instruction}

Personalized multimodal instruction focuses on guiding MLLMs to generate more tailored content through structured prompts and contextual signals. 
For example, CGSMP~\cite{yong2023cgsmp} demonstrates controllable text summarization using multimodal prompts based on image entities, reducing hallucinations and improving summarization quality.
\citet{li2024multimodal} further propose multimodal in-context tuning leveraging in-context learning abilities of MLLMs to dynamically generate product descriptions based on visual and textual cues. 

\subsection{Personalized Multimodal Alignment}\label{sec:text-gen-alignment}
To better reflect user intents in generated texts, a few works explore aligning multimodal inputs to personalized user preferences.
For instance, MPDialog~\cite{agrawal2023multimodal} aligns character personas and visual scenes to generate context-consistent dialogues.
Athena 3.0~\cite{fan2023athena} apply this concept to conversational agents, fusing neuro-symbolic strategies with multimodal dialogue generations, aligning responses with user preferences in dynamic contexts.
In addition, \citet{sugihara2024language} aligns video summarization with user-defined semantics by matching textual and visual content, ensuring personalized summaries.

\subsection{Personalized Multimodal Generation}\label{sec:text-gen-generation}

To generate text that aligns more closely with user-specific preferences,
\citet{wu2024understanding} introduce a framework for personalized video commenting, where clip selection and text generation processes are tailored to user preferences.
\citet{wang2024personalized} generate personalized time-synchronized comments on videos by leveraging a multimodal transformer to integrate visual elements with user-specific commentary.

\subsection{Personalized Multimodal Fine-tuning}\label{sec:text-finetune}
While prompting and instructions might not always achieve satisfying performance, several fine-tuning methods have been developed to help better adapt pre-trained MLLMs to specific user contexts and tasks.
\citet{wang2023user} propose prefix-tuning for personalized image captioning, reducing computation costs while retaining high-quality, user-specific outputs.
\citet{pi2024personalized} propose visual instruction tuning to address the limitations of generic MLLMs by enabling them to recognize individuals in images through visual instructions and generate personalized dialogues.

\section{Personalized MLLM Image Generation}\label{sec:image-gen}

\subsection{Personalized Multimodal Instruction}\label{sec:image-gen-instruction}
\citet{zhong2024user} propose a novel multimodal prompt to include complex user queries for customized instructions.
\citet{gal2022image} enables multimodal input to be tokenized into a lookup table, whose indexes are further used for generation based on a text transformer.
MuDI~\cite{jang2024identity} addresses identity mixing in multi-subject text-to-image personalization by leveraging segmented subjects using the Segment Anything Model~\cite{kirillov2023segment}. MuDI employs data augmentation (Seg-Mix) during training and an innovative inference initialization technique to generate distinct multi-subject images without mixing identities.
Subject-Diffusion~\cite{ma2024subject} introduces an open-domain personalized text-to-image generation framework that does not require test-time fine-tuning and relies on a single reference image for generating personalized images. The method combines text and image features using a custom prompt format, integrates fine-grained object features and location control for enhanced fidelity, and employs cross-attention map control to handle multiple subjects simultaneously.

\subsection{Personalized Multimodal Alignment}\label{sec:image-gen-alignment}
MoMA~\cite{song2024moma} is a tuning-free, open-vocabulary model for personalized image generation which combines reference image features with text prompts, enabling flexible re-contextualization and texture editing while preserving high detail fidelity and identity.
$\lambda$-ECLIPSE~\cite{patel2024lambda} leverages CLIP latent space to accelerate and facilitate personalized generation.

\subsection{Personalized Multimodal Generation}\label{sec:image-gen-generation}
\citet{kim2024layout} propose Layout-and-Retouch, an approach to achieve better diversity in the personalization of image generation.
\citet{shi2024instantbooth} propose Instantbooth for personalized generation without test-time fine-tuning. 

\subsection{Personalized Multimodal Fine-tuning}\label{sec:image-gen-finetune}
MS-Diffusion~\cite{wang2024ms} introduces a zero-shot, layout-guided method for multi-subject image personalization in diffusion models. It integrates a Grounding Resampler to enhance subject detail extraction and a Multi-Subject Cross-Attention mechanism to manage conflicts in multi-subject scenarios, ensuring accurate representation of subjects in specific regions while preserving overall image fidelity.
UNIMO-G~\cite{li2024unimo} unifies the multimodal end-to-end fine-tuning with visual and language transformer models.

\section{Personalized MLLM Recommendation}\label{sec:rec}

\subsection{Personalized Multimodal Instruction}\label{sec:rec-instruction}
Multimodal instructions in recommendations allow for the personalization of user intentions and preferences~\cite{ye2024harnessing,zhou2023exploring},
rich contexts of item information~\cite{zhou2023exploring,lyu2024x}, and more diverse user-system interactions~\cite{karra2024interarec}.
To better express preferences for novel items, the user can provide the reference image as part of the recommendation instruction~\cite{zhou2023exploring}.
Based on the user's interaction with visual items, the multimodal recommender system could analyze user preferences and intentions~\cite{ye2024harnessing,yu2024ra,liu2024rec}.
PMG~\cite{shen2024pmg} further transform multimodal user behaviors into language to model user preferences for a recommendation task.
For items in recommendation tasks, when meta data of a item is lacking, 
MLLMs are beneficial to extract rich descriptive information of items for better recommendation~\cite{zhou2023exploring,lyu2024x}.
On the other hand, many works also encode visual information with textual information into latent representations for better item modeling~\cite{tian2024mmrec,wei2024towards}.
In addition, multimodal instructions enable interactions through screenshots~\cite{karra2024interarec}, personalized item design~\cite{wei2024towards}, 
and conversational recommendations based on reference images~\cite{zhou2023exploring}.

\subsection{Personalized Multimodal Alignment}\label{sec:rec-alignment}
To unify the understanding and reasoning of multimodal information, 
MLLMs can serve to transcript visual information into texts~\cite{shen2024pmg,ye2024harnessing}, 
enable latent representation fusion and alignment~\cite{tian2024mmrec,xu2024align,hu2024lightweight},
and unify multimodal modeling~\cite{wei2024towards,liu2024rec,yu2024ra}.
Based on the user's previous interaction with visual items, 
the MLLMs can provide explanations on the interacted items' descriptions~\cite{ye2024harnessing},
and also the user's interaction behaviors~\cite{shen2024pmg},
which can be further used as part of textual prompts for further textual-only LLM-based reasoning.
On the other hand, other works consider using MLLMs' abilities in encoding multimodal representations,
to enable item-level augmentation~\cite{tian2024mmrec}, user-item fusion~\cite{xu2024align}, and across-task multimodal knowledge transferring~\cite{hu2024lightweight}.
Recent developments in end-to-end  multimodal learning, where multimodal instructions input as a sequence,
enables a novel paradigm of generative recommendation, which regards recommends as a next-token prediction task.
Some preliminary works directly prompt advanced MLLMs (\eg, GPT-4V) to understand the multimodal instruction and recommendations based on reasoning.
\citet{liu2024rec} designs such prompting methods in sequential recommendation tasks, whose recommendation results are re-ranked after generation.
Some works also enable tokenized items and users~\cite{yu2024ra} with multimodal information, which can directly generate items from the model's embedding space.

\subsection{Personalized Multimodal Generation}\label{sec:rec-generation}
Generative recommender systems leverage next-token generation as a unified recommendation policy.
The LLM can directly generate items as language tokens by further encoding items into the LLM's embedding space.
Recent ID-based representation learning methods encode item IDs into language embeddings,
learned from multimodal and collaborative knowledge~\cite{yu2024ra}.
In addition, some unified framework~\cite{wei2024towards} enables encoding of multi-channel information,
and recommendation generation as well as modified images of the user's potentially interested items.
Such multimodal generation provides better explanability of the recommended items and better convinces users to accept the items.
However, since item re-ranking is one of the essential steps for post-processing, how to leverage multimodal output for item re-ranking can be still under-explored.

\subsection{Personalized Multimodal Fine-tuning}\label{sec:rec-finetune}
To achieve more efficient alignment of personalized MLLM recommendations, 
several works also propose fine-tuning methods on MLLMs.
GPT4Rec~\cite{zhang2024gpt4rec} incorporates graph modality information which enables structure-level prompting.
Based on the novel prompt design, GPT4Rec performs prompt tuning to benefit streaming recommendations on both the node level and the view level.
InstructGraph~\cite{wang2024instructgraph} also leverages the graph structure to unify NLP, information retrieval, and recommendation tasks, 
and thus further enables fine-tuning and RLHF for alignment.
MMSSL~\cite{wei2023multi} is a unified learning framework to first decompose users' modality-aware preferences,
and then collaboratively learn the inter and inter-dependency and inter-modality preference signals through self-augmentation. 
\citet{deng2024end} further propose a unified transformer model that enables inputs of multimodal information,
and outputs of content features which can be used to pair with item representations for direct recommendation tasks.

\section{Personalized MLLM Retrieval}\label{sec:ret}

\subsection{Personalized Multimodal Instruction}\label{sec:ret-instruction}

Personalized multimodal instruction focuses on improving the ability of MLLMs to tailor their outputs based on user-specific needs and preferences.
Existing benchmarks, such as ConCon-Chi~\cite{Rosasco_2024_CVPR}, raise challenges in personalized text-to-image retrieval by introducing complex and varied contexts and instructions for personalized concept learning and compositionality assessment.
In a different direction, the Learnable Agent Collaboration Network~\cite{shi2024learnable} proposes a framework where multiple agents with distinct instructions and roles collaborate to deliver user-specific responses in multimodal search and retrieval engines.
Med-PMC~\cite{liu2024med} creates a simulated clinical environment where MLLMs are instructed to interact with a patient simulator decorated with personalized actors for multimodal information retrieval and decision making.
These works highlight the need for MLLMs to effectively integrate multimodal information and personalize their responses across diverse user interactions.

\subsection{Personalized Multimodal Alignment}\label{sec:ret-alignment}

To enhance the interaction between MLLMs and user-specific inputs, personalized multimodal alignment ensures that models can adapt to unique preferences and contexts.
AlignBot~\cite{chen2024alignbot} aligns robot task planning with user reminders by using a tailored LLaVA-7B model as an adapter for GPT-4o.
The alignment translates user instructions into structured prompts enabling a dynamic retrieval mechanism that recalls relevant past experiences and improves task execution.
In contrast, the Align and Retrieve framework~\cite{xu2024align} focuses on image retrieval with text feedback, using a composition-and-decomposition learning strategy to unify visual and textual inputs.
This approach creates a robust multimodal representation for precise alignment between composed queries and target images.
Both methods underscore the importance of aligning multimodal inputs for complex retrieval tasks with personalized user needs.

\subsection{Personalized Multimodal Generation}\label{sec:ret-generation}

Capturing personalized user intents for more accurate retrieval results is another challenge for MLLMs.
\citet{ye2024contemporary} propose an iterative user intent expansion framework, demonstrating how MLLMs can parse and compose personalized multimodal user inputs.
It refines the image search process through stages of parsing and logic generation, which also allows user to iteratively refine their search queries.
Similarly, \citet{wang2024multimodal} develop a multimodal query suggestion method leveraging multi-agent reinforcement learning to generate more personalized and diverse query suggestions based on user images, thereby improving the relevance of retrieval results.
Additionally, \citet{nguyen2024yo} present Yo'LLaVA, a personalized assistant that embeds user-specific visual concepts into latent tokens, enabling model tailored interactions and retrievals.
These methods collectively emphasize the integration of generation techniques into retrieval systems for more precise and personalized retrieval outcomes.

\subsection{Personalized Multimodal Fine-tuning}\label{sec:ret-finetune}

To further improve the retrieval capabilities of MLLMs in personalized contexts, various fine-tuning techniques are developed.
FedPAM~\cite{feng2024fedpam} introduces a federated learning approach for fine-tuning text-to-image retrieval models, allowing them to adapt to user-specific data without sharing confidential information, thereby addressing the data heterogeneity challenge.
VITR~\cite{gong2023vitr} enhances vision transformers for cross-modal information retrieval by refining their ability to understand relationships between image regions and textual descriptions.
\citet{Yeh_2023_CVPR} further demonstrate how models can be adapted to identify specific user-defined instances, such as objects or individuals in videos, by extending the model's vocabulary with learned instance-specific features.
Additionally, \citet{chen-etal-2023-task} explore task-personalized fine-tuning for visually-rich document entity retrieval, utilizing meta-learning to extract unique entity types with few examples.
Furthermore, \citet{li2024generative} propose a generative cross-modal retrieval framework that fine-tunes MLLMs to memorize and retrieval visual information directly from model parameters, offering a novel approach to image retrieval.
These works show great potential of fine-tuning MLLMs to enhance their retrieval performance in personalized and diverse multimodal tasks.

\section{Evaluation}\label{sec:eval}
The evaluation of personalized MLLMs is typically categorized based on the target task. UniMP~\cite{Wei2024} explores various personalized tasks, such as personalized preference prediction, personalized explanation generation, and user-guided image generation, among others. 
Several models focus on personalized recommendation tasks, as detailed in Section~\ref{sec:rec}~\cite{karra2024interarec, wei2023multi, zhang2024gpt4rec, ye2024harnessing}. In the recommendation setting, the goal is to rank the true target (\eg, item or movie) highest on the list relative to other items. Commonly used metrics for this task include MRR, Recall@k, Hit@k, AUC, HR@k, and NDCG@k, which evaluate how well the model ranks the true target item in comparison to other options.

Personalized multimodal generation focuses on creating customized content, such as images or text, by considering user-specific behavior. This includes generating personalized images, posters for movies, or emojis, as demonstrated by \citet{shen2024pmg}. \citet{shen2024pmg} utilize various image similarity techniques to evaluate the similarity between the generated content and historical or target items, employing metrics like LPIPS (Learned Perceptual Image Patch Similarity)~\cite{zhang2018unreasonable} and SSIM (Structural Similarity Index Measure)~\cite{wang2004image}. Additionally, this area encompasses personalized chatbots~\cite{nguyen2024yo, fei2024empathyear, abuzuraiq2024towards}, which assess models' abilities to recognize personalized subjects in images, handle visual and text-based question answering~\cite{alaluf2024myvlm, nguyen2024yo}, and evaluate emotional intelligence by measuring emotion detection accuracy and response diversity~\cite{fei2024empathyear}.

\citet{gal2022image} introduce personalized text-to-image generation, synthesizing novel scenes based on user-provided concepts and natural language instructions. They evaluate the model by calculating the average pair-wise CLIP-space cosine similarity between generated images and the concept-specific training set, as well as the editability of prompts by measuring the similarity between the generated images and their textual descriptions using CLIP embeddings. Other methods in this domain~\cite{kim2024layout, song2024moma} focus on two main aspects: identity preservation, which assesses the model's ability to maintain the subject's identity, and prompt fidelity, which ensures alignment between the generated images and the textual prompts. Identity preservation is typically measured by I-CLIP~\cite{radford2021learning}  and I-DINO~\cite{caron2021emerging}, which compute subject similarity using CLIP and DINO as backbones. Prompt fidelity is evaluated through the CLIP-based text-image similarity score (T-CLIP). Image diversity is assessed using the Inception Score (IS) to capture the variation within generated sets. \citet{jang2024identity} introduce a new metric, Detect-and-Compare (D\&C), to evaluate multi-subject fidelity, addressing the limitations of existing metrics (like I-CLIP or DINOv2) that do not to prevent identity mixing in multi-subject scenarios. \citet{wang2024ms} and others~\cite{ma2024subject, li2024unimo} focus on multi-subject personalized text-to-image generation, using M-DINO to capture subject fidelity by avoiding subject neglect, which average fidelity metrics may overlook.

Other tasks include personalized image retrieval, where the Vision-Language model is expected to retrieve a collection of relevant images based on a textual query, using personalized context previously provided by the user (either in the form of images or text). \citet{cohen2022my} first introduce the concept of Personalized Vision and Language, along with a benchmark to evaluate models on tasks like personalized image retrieval and personalized image segmentation. ConCon-Chi~\cite{Rosasco_2024_CVPR} further extend this by proposing a new benchmark that evaluates models' ability to learn new meanings and their compositionality with known concepts. The setting of personalized retrieval has also been expanded to videos in the works of~\cite{korbar2022personalised,yeh2023meta}. Zero-shot Composed Image Retrieval (ZS-CIR) evaluates the model's capability to retrieve images based on compositional queries, without requiring prior examples for new combinations of known concepts. The metrics typically used for these tasks include measuring the rank of the first ground truth (GT) image using Mean Reciprocal Rank (MRR), Recall@k to determine if any GT image appears in the top-k results, and Mean Average Precision (MAP) to assess the ranking of all GT images. Additionally, MAP@k evaluates the precision of GT images up to the top-k retrieved results.
Lastly, personalized semantic segmentation focuses on segmenting an instance of a personalized concept in an image, based on a textual query that refers to that concept. \citet{cohen2022my} use the intersection-over-union (IoU) metric to evaluate this, reporting the rate of predictions with IoU above a specified threshold.

\section{Datasets}~\label{sec:datasets}

In recent years, the field of multimodal and personalized learning has seen an increase in datasets, each designed to address specific research challenges. These datasets span various domains, including vision-language models, agent collaboration networks, fashion retrieval, and cross-modal retrieval tasks. For instance, ConCon-Chi~\cite{Rosasco_2024_CVPR} and MSMTPInfo~\cite{shi2024learnable} provide benchmarks for evaluating complex, dynamic user interactions in multimodal contexts, while fashion-focused datasets such as FashionIQ~\cite{wu2021fashion}, and Fashion200k~\cite{han2017automatic} offer rich collections of images and triplets for advancing research in fashion retrieval and recommendation.
UniMP~\cite{Wei2024} uses Amazon review data including the user-item interactions and the items' images. Other datasets like RefCOCOg~\cite{7780378} and CLEVR~\cite{johnson2017clevr} focus on relationships between objects and regions in images, contributing to cross-modal retrieval research.
In addition, a wide range of datasets in multimodal recommendation and information retrieval are used in developing and evaluating personalized MLLMs.
We summarize a  series of comprehensive datasets with detailed descriptions in Table~\ref{table:datasets} (Appendix~\ref{sec:app-data}).

\section{Open Problems \& Challenges }\label{sec:open-problems-challenges}
In this section, we discuss open problems and highlight important challenges for future work.

\subsection{Benchmark Datasets}
For developing better personalized MLLMs, there is a need for more robust and  comprehensive benchmark datasets to improve both training and evaluation.
Currently, there are limited multimodal benchmark datasets with user-specific information.

\subsection{Evaluation Metrics}
Many works have focused mainly on evaluating downstream tasks such as recommendation, rather than directly assessing the quality of generated outputs. However, direct evaluation of generation quality is crucial for improving these models.

\subsection{Multimodality Diversity and Complexity}
Most existing work leverages only standard modalities such as text and images.
Future work should explore other more diverse types of modalities such as audio, video, graphs, among others.
Furthermore, there is a need to develop techniques for supporting many more modalities all at once, as most work has focused only on two such modalities.

\subsection{Modality Fusion}
In MLLMs, a common challenge is the dominance of text during modality fusion. Since these models are typically pre-trained on vast amounts of text data, they become highly proficient at processing and interpreting textual information. Consequently, when integrating multiple modalities, there is a tendency for the model to over-rely on text, which can overshadow other crucial data sources like images or audio. This text bias often results in suboptimal performance in tasks that require a deeper understanding of non-textual information, where visual or audio cues are key to provide the full context.

\subsection{Theoretical Foundations}
Developing theoretical foundations for the techniques behind personalized MLLMs remains an open problem~\cite{wu2024commit}. 
Understanding their theoretical limits and trade-offs is also of fundamental importance.
In spite of this, understanding the theoretical limits of these techniques remains an open problem for future work.

\section{Conclusion}\label{sec:conc}
In this work, we present a comprehensive survey on personalized multimodal large language models, focusing on their architectures, training methods, and applications. We introduce an intuitive taxonomy for categorizing the techniques used to personalize MLLMs for individual users and provide a detailed discussion of these approaches.
Additionally, we explore how these techniques can be combined or adapted when appropriate, highlighting both their advantages and underlying principles. We offer a concise summary of the personalization tasks addressed in existing research and review the evaluation metrics used for each task.
We also summarize key datasets that are valuable for benchmarking personalized MLLMs. Finally, we identify important open challenges that remain to be addressed.
This survey serves as a valuable resource for researchers and practitioners seeking to understand and advance the development of personalized multimodal large language models.

\section*{Limitations}
In this paper, the extent of personalization in MLLMs is inherently limited by the available datasets and applications. 
Moreover,
our focus is on MLLMs that incorporate specific personalized multimodal instructions, but we do not address inherent model biases that may affect personalization. Addressing such biases could be a valuable direction for future research in MLLM personalization.

\bibliography{main}
\bibliographystyle{acl_natbib}

\appendix

\section{Summary of Datasets in Personalized MLLM Recommendation and Retrieval}\label{sec:app-data}

\begin{table*}[ht]
\centering
\resizebox{\textwidth}{!}{
\begin{tabular}{p{6cm}p{8cm}p{8cm}}
\toprule
\textbf{Dataset Name} & \textbf{Description} & \textbf{Stats} \\
\midrule
ConCon-Chi \cite{Rosasco_2024_CVPR} &  Benchmark for personalized vision-language tasks & 4008 images; 20 concepts; 101 contexts \\
\midrule
MSMTPInfo \cite{shi2024learnable} & Evaluation for Agent Collaboration Network & 13 themes; multiple sessions; dynamic user topics \\
\midrule
Shoes \cite{10.5555/1886063.1886114} & Dataset for interactive fashion image retrieval & 8900 training triplets; 1700 test triplets \\
\midrule
Fashion200k \cite{han2017automatic} & Large-scale fashion dataset with over 200K images & 172K training images; 33K test queries \\
\midrule
Business Dataset \cite{5206848} & User query images from a real image search engine & 50K images; 5 suggestions per image \\
\midrule
Yo'LLaVA \cite{nguyen2024yo} & Personalized language and vision assistant dataset & 40 subjects; 10-20 images per subject
 \\
\midrule
RefCOCOg \cite{7780378} & Images from MS-COCO with relational annotations & 21899 train, 1300 val and 2600 test images \\
\midrule
% CLEVR  \cite{johnson2017clevr}& Images depicting 3D-rendered objects with relational annotations & 30000 train, 1000 validation and 1000 test images \\
% \midrule
Flickr30K \cite{young-etal-2014-image}& Benchmark for visual-semantic embedding networks & 29000 train, 1000 validation and 1000 test images \\
\midrule
MicroLens \cite{ni2023content} & Video introductions and cover images & 1 billion user-item interactions; 34 million users \\
\midrule
Amazon-Baby \cite{mcauley2015image} & Images and product descriptions & 180 million relationships; 6 million objects; \\
\midrule
interaRec \cite{karra2024interarec} & Screenshot based recommendations using multimodal large language models & 1,500 sessions; item IDs; screenshots of webpages \\
\midrule
POG \cite{chen2019pog} & Fashion rec by personalized outfit generation & 1.43 million outfits; 80 most frequent categories; \\
\midrule
MovieLens \cite{ni2023content} & Describes 5-star rating and free-text tagging activity & 100K ratings; 3,683 tag applications; 9,742 movies \\
\bottomrule
\end{tabular}
}
\caption{Summary of Datasets.}
\label{table:datasets}
\end{table*}

To tackle these diverse challenges, researchers employ a wide array of approaches, ranging from the use of established, large-scale public datasets to the creation of tailored datasets for specific tasks. For example, studies on multimodal sequential recommendation often leverage datasets like Amazon-Baby~\cite{mcauley2015image}, Amazon-Game~\cite{he2016ups}, and MicroLens~\cite{ni2023content} for evaluation. In contrast, some researchers harness the power of Large Language Models (LLMs) to generate synthetic data, as exemplified by the Business Dataset~\cite{5206848}, which utilizes GPT-generated suggestions for labeling. The development of custom datasets also plays a crucial role, as seen in POG's~\cite{chen2019pog} adaptation of the iFashion dataset for personalized fashion recommendations, InteraRec's~\cite{karra2024interarec} collection of screenshots from Amazon websites to create a new resource for multimodal recommendation research, and LongLaMP~\cite{kumar2024longlamp} benchmark to evaluate long-form personalized text generation. A summary of these datasets can be found in Table~\ref{table:datasets}.
 
\section{Applications}\label{sec:applications}
Personalized MLLMs have an extensive range of applications, targeting various tasks in the textual, visual, audio, and other domains.

\subsection{Personalized MLLM Recommendation} 
\citet{liu2024beyond} develop a multimodal knowledge graph that recommends missing entities in triplet structures.
Their approach predicts relationships between entities (\eg, people) within images.

\subsection{Personalized MLLM Retrieval}
\citet{choudhury2024etdpc} classify Electronic Theses and Dissertations (ETD) through a combination of visual and textual learning.

\subsection{Personalized MLLM Text Generation}
\citet{wei2024towards} propose a multimodal learning framework where a vision model extracts features from images and a language model learns from texts.
The extracted features are jointly modeled to yield personalized product recommendation, preference prediction, explanation generation.
Additionally, \citet{wang2024t} leverage multimodal learning to answer science-related questions using chain-of-thought reasoning.

\subsection{Personalized MLLM Image Generation}
\citet{shen2024pmg} utilize MLLMs to generate movie posters tailored to users' preferences.
Similarly, ~\citet{song2024moma} and ~\citet{wei2024mm} leverage LLMs to generate images based on visual and textual prompts.

\subsection{Miscellaneous Applications}
MLLMs have been applied in various other fields, such as helping visually impaired individuals by verifying images and offering outfit suggestions~\cite{xie2024emerging}.
Moreover, MLLMs are used on brain tumor segmentation and tumor identification~\cite{dai2024federated}.

\end{document}